\documentclass{article}
\usepackage{spconf,amsmath,graphicx}
\usepackage{amsfonts,amssymb}
\usepackage{booktabs}
\usepackage{multirow}
\usepackage{amsmath}
\usepackage{amsfonts,amssymb}
\usepackage{color}
\title{Learning Subjective Time-Series Data via Utopia Label Distribution Approximation}
\name{ Wenxin Xu$^{a\star}$ \quad Hexin Jiang$^{a\star}$ \quad Xuefeng Liang$^{a}$  \quad Ying Zhou$^{a}$ \quad  Yin Zhao$^{b}$  \quad  Jie Zhang$^{b}$  \thanks{$^{\star}$Both authors contributed equally to this research.}}
\address{$^{a}$ School of Artificial Intelligence, Xidian University, China. \qquad $^{b}$ Alibaba Group, China. \\
xliang@xidian.edu.cn}
\begin{document}
%\ninept
%
\maketitle
\begin{abstract}
Subjective time-series regression (STR) tasks have gained increasing attention recently. However, most existing methods overlook the label distribution bias in STR data, which results in biased models. Emerging studies on imbalanced regression tasks, such as age estimation and depth estimation, hypothesize that the prior label distribution of the dataset is uniform. However, we observe that the label distributions of training and test sets in STR tasks are likely to be neither uniform nor identical. This distinct feature calls for new approaches that estimate more reasonable distributions to train a fair model. In this work, we propose Utopia Label Distribution Approximation (ULDA) for time-series data, which makes the training label distribution closer to real-world but unknown (utopia) label distribution. This would enhance the model's fairness. Specifically, ULDA first convolves the training label distribution by a Gaussian kernel. After convolution, the required sample quantity at each regression label may change. We further devise the Time-slice Normal Sampling (TNS) to generate new samples when the required sample quantity is greater than the initial sample quantity, and the Convolutional Weighted Loss (CWL) to lower the sample weight when the required sample quantity is less than the initial quantity. These two modules not only assist the model training on the approximated utopia label distribution, but also maintain the sample continuity in temporal context space. To the best of our knowledge, ULDA is the first method to address the label distribution bias in time-series data. Extensive experiments demonstrate that ULDA lifts the state-of-the-art performance on two STR tasks and three benchmark datasets.
\end{abstract}


%  \label{fig:teaser}\end{teaserfigure}


\maketitle

\section{Introduction}
With the quick development of multimedia social networks, many applications (e.g. audio-visual content-understanding \cite{naphade2002extracting}, movie-quality assessment \cite{lee2017consumption}, and personalized recommendations \cite{wei2019mmgcn}) require high-performance algorithms on tasks such as video affective analysis \cite{baveye2015liris}, video summarization \cite{gygli2014creating}, and music emotion recognition \cite{zhang2018pmemo}, to name a few. These tasks all focus on predicting subjective regression labels on long time-series data, which are named Subjective Time-series Regression (STR) tasks. Unlike other regression tasks (estimating age \cite{rothe2018deep}, depth \cite{silberman2012indoor} or health score \cite{quan1997sleep}), the data in STR task have two distinct properties: (1) their regression labels are subjective because labels are based on the subjective cognition of a group of people; and (2) they are continuous in both temporal context space and label space.

\begin{figure}
    \centering
    \includegraphics[width=75mm]{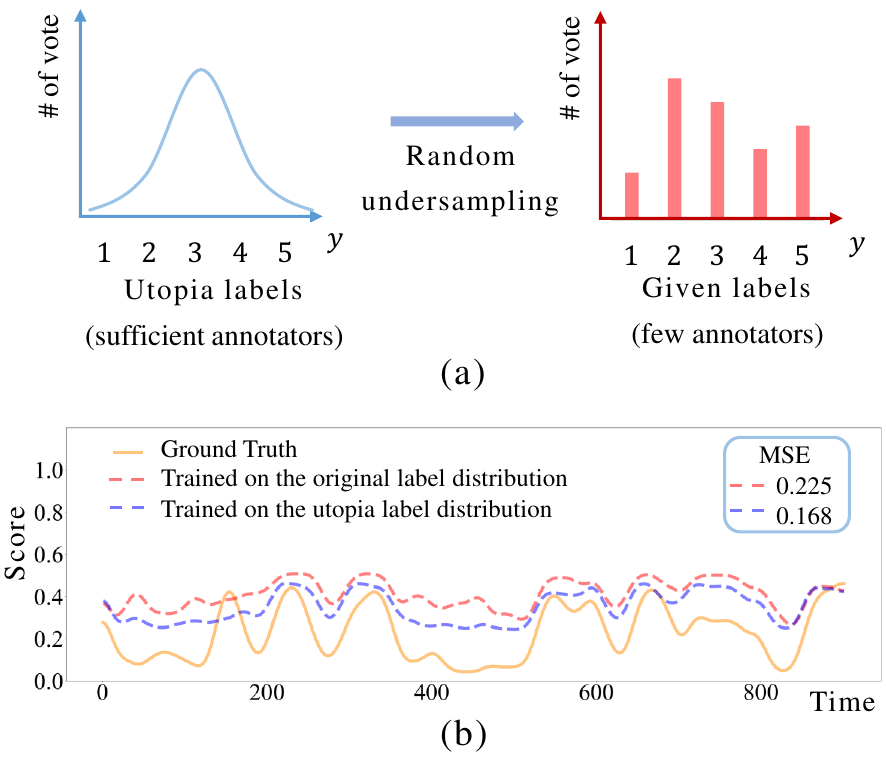}
    % \vspace{-0.5cm}
    \caption{(a) The utopia labels and the labels given by 20 annotators of a sample in the TVSum dataset \cite{song2015tvsum}. The label distributions in existing databases are likely biased due to insufficient annotators. (b) The comparison of predicting the importance score of each frame in a movie from the TVSum dataset. The dashed lines denote the results of PGL-SUM \cite{apostolidis2021combining} trained on the original label distribution given by the dataset (in red) and the utopia label distribution approximating real-world distribution (in blue), respectively. The ground truth is in yellow.}
    % \vspace{-0.5cm}
    \label{fig:utopia}
\end{figure}

Due to the specific nature of STR tasks, the regression label for each frame of data in datasets is commonly the average of annotators’ votes. We observe that most existing datasets have insufficient annotators (e.g. 3 persons for each annotation in the LIRIS-ACCEDE dataset \cite{baveye2015liris}, 15 $ \sim $  18 annotators in the SumMe  \cite{gygli2014creating} dataset, and 20 annotators in the TVSum \cite{song2015tvsum} dataset). In this case, samples and their labels in the training and test sets can be considered as a form of undersampling from the real world. It may result in a skewed label for each training sample, as shown in Fig. \ref{fig:utopia}(a), and further cause the whole training label distribution to be biased compared to the one in the real world. Moreover, the label distributions of training and test sets are likely to be neither identical nor uniform. This would make the traditional data balancing methods, which assume the training and test sets have the same distribution, inapplicable in this case. In a pilot study on the TVSum dataset, we found that a model trained on the original label distribution given by the dataset underperforms the one trained on the utopia label distribution, which approximates real-world distribution, in a task of predicting the importance score of each frame in a video. The result shown in Fig. \ref{fig:utopia}(b) demonstrates that the regression labels given by a few annotators are likely biased and degrade model performance.

The label distribution bias in regression tasks has recently received increased attention \cite{branco2017smogn,torgo2013smote, gong2022ranksim, ren2022balanced, steininger2021density, yang2021delving}. These studies hypothesize the training and test sets have the same distribution and the test set is uniformly distributed, then address this issue by balancing data distributions. They are primarily divided into two categories: \textit{Data-level methods}, which oversample or undersample subsets of training data to balance the distribution, such as SmoteR \cite{torgo2013smote}, SmogN \cite{branco2017smogn}, etc.; and \textit{Algorithm-level methods}, which tailor existing learning algorithms, particularly the loss functions, to cope better with imbalanced training data, such as BMC \cite{ren2022balanced}, Dense loss \cite{steininger2021density}, RankSim \cite{gong2022ranksim}, etc. An emerging work, DIR \cite{yang2021delving}, combines these two approaches.% All the above methods assume that the test data are balanced.

For STR tasks, the test data is sampled from the real world. Its label distribution is unknown and usually imbalanced. We think making the label distributions of the training and test sets more similar can improve the prediction accuracy of the model. Thus, a fair model could be obtained by training on a utopia label distribution, which approximates real-world distribution. To this end, we propose a method named Utopia Label Distribution Approximation (ULDA). The core of ULDA is based on the central limit theorem (CLT) \cite{kwak2017central}, which states the distribution of a sample variable (the label in this study) approximates a Gaussian distribution as the number of annotators becomes sufficiently large. Specifically, we convolve the original training label distribution by a Gaussian kernel to approximate the utopia label distribution. After convolution, the required sample quantity at each regression label may change. To tackle this issue, we devise two approaches: 1). When the required sample quantity after convolution is greater than the original sample quantity, we devise a Time-slice Normal Sampling (TNS) to augment the training samples. TNS estimates the normal distribution of sample features within a short time slice and uses Monte Carlo sampling to generate new samples that can maintain the contextual continuity. 2). When the required sample quantity is less than the original sample quantity, we devise a Convolution Weighted Loss (CWL) to lower the weights of these samples without undersampling. The weights are calculated using the ratio of sample quantities before and after convolution. Again, CWL can maintain the contextual continuity because no training samples are eliminated.

The contributions of this work are threefold:

1.	We investigate the nature of data distribution in STR tasks, and then propose the Utopia Label Distribution Approximation (ULDA) method to address the label distribution bias in time-series data. To the best of our knowledge, ULDA is the first method in the literature.
%, which approximates real-world distribution by convolving the original label distribution in datasets.

2.	We devise the Time-slice Normal Sampling (TNS), a data-level method for generating new time-series data when the required sample quantity is greater than the original sample quantity, and the Convolutional Weighted Loss (CWL), an algorithm-level approach for lowering the sample weight when the required sample quantity is less than the original sample quantity.

3.	Extensive experiments demonstrate that our proposed ULDA (TNS+CWL) considerably improves baseline models and outperforms the state-of-the-art (SOTA) methods on three benchmark datasets.

\section{Related Work}
\subsection{Subjective Time-series Regression Task}
As STR tasks have received increased attention, a number of datasets and approaches for STR tasks have recently been proposed. Zhang et al. \cite{zhang2018pmemo} created the PMEmo dataset for Music Emotion Recognition (MER) research, which contains emotion annotations of songs at each 0.5-second. Baveye et al. \cite{baveye2015liris} created the LIRIS-ACCEDE dataset, a large video dataset for affective content analysis that every second of videos is labeled. They also used an SVR model to test the feasibility of emotional regression prediction. Gygli et al. \cite{gygli2014creating} and Song et al.  \cite{song2015tvsum} built SumMe and TVSum video summarization datasets, respectively, in which each frame is assigned an importance score.

For STR tasks, Zhao et al. \cite{zhao2019video} designed a two-time-scale structure for movie affective analysis, which can capture both inter-clip and intra-clip relationships in order to utilize temporal context in videos. To better integrate multimodal features, they applied a residual-based progressive training strategy. Mittal et al. \cite{mittal2021affect2mm} proposed the Affect2MM for video affective analysis, which extracts video emotional features using an LSTM model. They reported Affect2MM was the first method that explicitly models temporal causality using attention and Granger causality. To address the weaknesses of previous approaches to modeling long-term dependencies, Apostolidis et al. \cite{apostolidis2021combining} introduced the PGL-SUM for video summarization, which models the dependencies of frames at different levels of granularity by combining global and local multi-head attention. Zhang et al. \cite{zhang2022enlarging} proposed the RMN, a reinforcement learning based memory network that alleviates the storage limitations of LSTM and the gradient vanishing/exploding problem for long sequence predictions. It is currently reported to be SOTA on a variety of STR tasks.

All about methods focused on learning better contextual features of time-series data. Unfortunately, they overlooked the label distribution bias caused by insufficient annotators in existing STR data, resulting in biased models.

\subsection{Imbalanced regression}
The imbalanced classification has been extensively studied \cite{wu2022fast, liu2019large, shu2019meta, yang2020rethinking, Kang2020Decoupling}. In contrast, the imbalanced regression has only recently attracted researchers’ attention. As the continuity in label space makes imbalanced regression different from imbalanced classification, researchers proposed either new data-level or new algorithm-level approaches to improve model performance while learning from imbalanced training sets.

Data-level methods are based on various sampling strategies. In general, they attempt to build a balanced training set by synthesizing new samples for pre-defined rare label regions and downsampling for normal label regions during data pre-processing. SmoteR \cite{torgo2013smote} proposed by Torgo et al. is based on the SMOTE \cite{chawla2002smote} and synthesizes new data by linearly interpolating both inputs and targets at rare regions. Branco et al. \cite{branco2017smogn, pmlr-v94-branco18a} proposed SmogN, which combines SmoteR with oversampling via Gaussian noise, meanwhile, devised the REBAGG algorithm, which ensembles regressors trained by different resampling methods.

Algorithm-level methods assume that the test sets are balanced and uniform, and then devise different reweighting losses to learn balanced features from imbalanced training sets. Yang et al. \cite{yang2021delving} and Steininger et al. \cite{steininger2021density} proposed LDS and KDE to take the continuity of labels into account. They first estimate the label density distribution using kernel density estimation, and then optimize models by weighted losses that are inversely proportional to the label density. Gong et al. \cite{gong2022ranksim} introduced RankSim, a ranking loss that takes into account both nearby and distant label relationships to learn a better continuous feature space. Zhang et al.  \cite{ren2022balanced} proposed the Balanced MSE that improves the MSE loss by preventing the underestimation of rare labels when the training set is imbalanced.

The aforementioned methods either explicitly or implicitly assume that the test data is distributed uniformly. Furthermore, sampling methods in non-temporal regression tasks may fail to maintain the contextual continuity of time-series data. Thus, these methods are infeasible for STR tasks. Instead, our ULDA (TNS+CWL) approximates the utopia label distribution while maintaining the continuity of time-series data in both label and temporal context spaces.

\section{Proposed Method}
It is well-known that learning on a biased data distribution results in a biased model. To have a fair model and better generalization, the model is expected to be trained on real-world data distribution. Therefore, our goal is to let models learn on the utopia label distribution, which approximates real-world distribution based on the original label distribution in the dataset.
\subsection{Problem Setting and Motivation}
In STR tasks, let $\{(x_i, y_i )\}_{i=1}^m$ be a time-series data with $m$ frames in the training set, where $x_i$ is an input sample and $y_i$ is the label, which is a continuous target. Assume the label space of the dataset has $b$ regression labels and the label $y$ has $n_y$ samples.

Due to few annotators in datasets, we consider that samples and their labels in the training and test sets are randomly undersampled from the real world. It may result in a skewed label for each training sample, and further cause the whole training label distribution to be biased. In probability theory, the central limit theorem\cite{kwak2017central} tells us that the label mean tends towards a Gaussian distribution when the number of annotators is sufficiently large in the real world. Given $n$ samples with the true label $y$, their actual labels will follow a Gaussian distribution $N(y, \sigma_y^2)$,
\begin{equation}
%\begin{array}{c}
    \mu_{y}=\frac{\sum_{i=1}^n y}{n}=y,\qquad  \sigma_{y}=\frac{\sum_{i=1}^n\sigma_i^2}{n},
%\end{array}
\end{equation}
where $\sigma_i^2$ denotes the variance of people’s inconsistent votes. A detailed theoretical analysis can be found in the supplementary material.
%\begin{equation}
%\sigma_{y}=\frac{\sum_{i=1}^{q_{y}}\sigma_i^2}{q^2_{y}.}
%\end{equation}

We can see that, fortunately, approximating the utopia label distribution equates to Gaussian kernel regression in label space. This inspires us to propose the Utopia Label Distribution Approximation. It consists of three parts: 1) Label Distribution Convolution, which approximates the utopia label distribution. 2) Time-slice Normal Sampling (TNS), which is a data-level approach. 3) Convolution Weighted Loss (CWL), which is an algorithm-level approach. The details are given in below.

\subsection{Label Distribution Convolution}

\begin{figure}
    \centering
    \includegraphics[width=80mm]{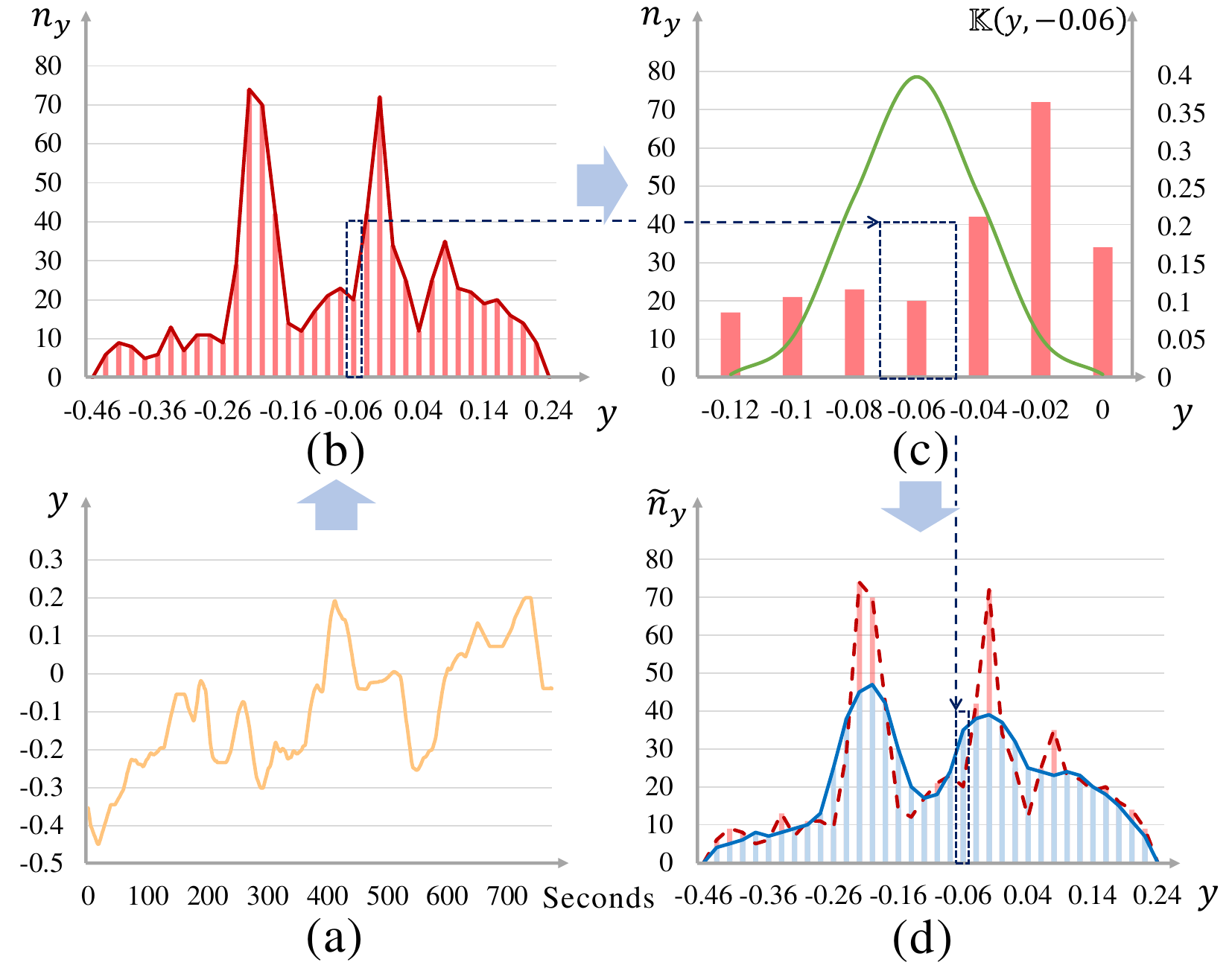}
    % \vspace{-0.5 cm}
    \caption{The procedure of approximating the utopia label distribution. (a) Regression labels of a movie along the timeline, which is from LIRIS-ACCEDE. (b) The original label distribution (in red) obtained from label statistics of the movie. (c) A Gaussian convolution at the regression label -0.06. The green curve is the Gaussian kernel. (d) The approximated utopia label distribution (in blue) after convolution.}
    \label{fig:method}
\end{figure}

Figure \ref{fig:method} shows the procedure of approximating the utopia label distribution. The step from Fig. \ref{fig:method}(a) to Fig. \ref{fig:method}(b) is label statistics that is carried out on each time-series data (video, music, etc.) in the training set. Let $n_y$ denote the sample quantity at the label $y$. Then, the original label distribution can be seen in Fig. \ref{fig:method}(b). Afterwards, a symmetric Gaussian kernel $\mathbb{K}(\cdot)$ is employed to convolve each regression label as shown in Fig. \ref{fig:method}(c),
\begin{equation}
    \Tilde{n}_{{y}'}= \int_{{y}'-\frac{\delta}{2}}^{{y}'+\frac{\delta}{2}} \mathbb{K}(y,{y}')n_ydy,
\end{equation}
where $\delta$ is the kernel size, $n_{y}$ and $\Tilde{n}_{{y}'}$ are the sample quantities of labels $y$ and ${y}'$ before and after convolution, respectively. Fig. \ref{fig:method}(d) shows the convoluted label distribution that is getting closer to the utopia label distribution and becomes smoother.

\subsection{Time-slice Normal Sampling}
After convolution, we can observe that the required sample quantity at some regression labels is greater than the original one, as shown in Fig. \ref{fig:method}(d). SmoteR \cite{torgo2013smote} and SmogN \cite{branco2017smogn} addressed this problem by oversampling the original samples. However, the continuity in both label and temporal context spaces makes samples in STR tasks different from samples in the age and depth estimation tasks \cite{yang2021delving,ren2022balanced,gong2022ranksim}. For example, samples with the same label may locate at different timepoints in a time-series data. They may have significantly different appearances, such as the scene change and viewpoint switching, as shown in Fig. \ref{fig: TNS}(a). Thus, conventional methods (e.g. Mixup \cite{zhang2017mixup} and SMOTE \cite{chawla2002smote} that generate new samples by randomly combining original samples) may fail to maintain the continuity of samples in temporal context space.

To address this issue, we propose Time-slice Normal Sampling (TNS) using neighboring samples to maintain the contextual continuity. TNS  estimates the feature distribution of local samples in a short time slice using a normal distribution, then augments samples by Monte Carlo sampling on the estimated distribution. The detailed steps are as follows:

\begin{figure}
    \centering
    \includegraphics[width= 1\linewidth]{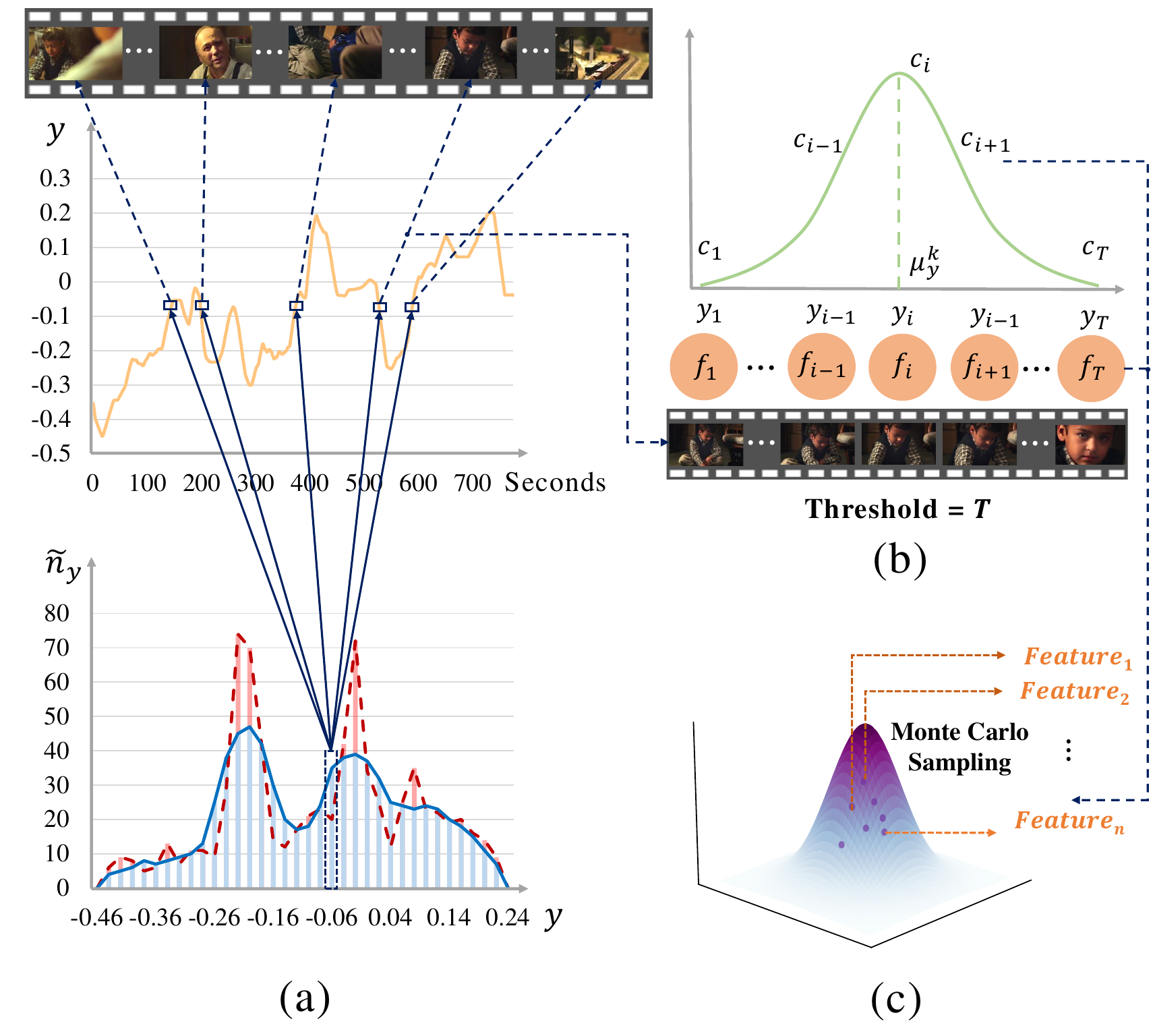}
    % \vspace{-0.5 cm}
    \caption{(a) The label distribution of a movie in LIRIS-ACCEDE before and after convolution. The label -0.06 requires new samples after convolution. Its original samples locate at different timepoints and have very different appearances. (b) Samples with nearly identical and continuous labels in a slice. The feature normal distribution is estimated according to the contribution of these samples. (c) Monte Carlo sampling from the feature normal distribution to generate new samples in feature space.}
    % \vspace{-0.5 cm}
    \label{fig: TNS}
\end{figure}

\textbf{3.3.1 New Sample Allocation:} First, we need to determine the appropriate timepoints to which the new samples should be allocated. Assume these samples are distributed across $z$ different slices in the time-series data, denoted as $\{s_{y}^1, s_{y}^2,…,s_{y}^z\}$. Their corresponding slice lengths are $\{l_{y}^1, l_{y}^2,…,l_{y}^z\}$. $l_{y}^k,(k=1,…,z)$ is the number of frames whose labels are nearly identical and continuous. The probability of selecting the slice $s_{y}^k$ to allocate new samples is proportional to its length,
\begin{equation}
    P(s_{y}^k )=\frac{l_{y}^k}{\sum_{k=1}^z l_{y}^k}.
\end{equation}
This ensures that new samples are added to the time-series data evenly along the timeline.

\textbf{3.3.2 Normal Distribution Estimation for Slice Features:} After determining the slice $s_{y}^k$, oversampling will be carried out on the original samples in $s_{y}^k$. Since the normal distribution is suitable for modeling continuous variables in temporal context space and has good mathematical properties, the sample augmentation is implemented by sampling from an estimated distribution in feature space in this work. We first estimate the normal distribution for the feature set of original samples in $s_{y}^k$. $F^k_{y} \in {R^{l_{y}^k \times d}}$  denotes the feature set, where $d$ is the feature dimension. To avoid the problem of inaccurate normal distribution estimation due to a small sample quantity in $s_{y}^k$, we set a threshold $T$. If $l_{y}^k \geqslant T$, the normal distribution is estimated using only $F^k_{y}$. If $l_{y}^k < T$, $s_{y}^k$ is extended in both directions until its length reaches $T$. Then, the normal distribution is estimated using the extended feature set $\hat{F}^k_{y}$.

Since the extended slice includes some samples with other labels, whose features may greatly differ from those of the original ones, they cause the estimated normal distribution to be shifted. To tackle this problem, we take into account the contribution of each sample feature to the distribution estimation. Specifically, the label mean $\mu_{y}^k$ of the original feature set $F^k_{y}$ and the label standard deviation $\sigma _{y}^k$ of the extended feature set $\hat{F}^k_{y}$ are used as prior to estimate a Gaussian-like probability density function of labels
\begin{equation}
    p\left( y_{i} \right) = \frac{1}{\sqrt{2\pi}\sigma^k_{y}}e^{- \frac{1}{2}{\left(\frac{y_{i} - \mu^k_{y}}{\sigma^k_{y}}\right)}^{2}}.
\end{equation}
The normalized contribution/weight of each sample feature $f_i$ in the slice, as shown in Fig. \ref{fig: TNS}(b), is
\begin{equation}
    c_{i} = \frac{p( y_i)}{\sum_{j = 1}^{T}{p(y_j)}}.
\end{equation}
When the label of a sample feature is closer to $\mu_{f_y^k}$, it contributes more to estimating the distribution, and vice versa.
Now, a more reasonable normal distribution $N(\mu_{f_y^k},\Sigma_{f_y^k})$ can be estimated by the weighted mean and weighted covariance matrix. They are calculated as follows:
\begin{equation}
    \mu_{f_y^k} = {\sum_{i = 1}^{T}{c_{i}f_{i}}}, \qquad \Sigma_{f_y^k} = \sum_{i = 1}^{T}{c_i\left( {f_{i} - \mu_{f_y^k}} \right)\left( {f_{i} - \mu_{f_y^k}} \right)^{\top}}. % \in \mathbb{R}^{d \times d}}. % \in \mathbb{R}^{d},
\end{equation}
%\begin{equation}
%    \Sigma_{f_y^k} = \sum_{i = 1}^{l^k_{y}}{c_i\left( {f_{i} - %\mu_{f_y^k}} \right)\left( {f_{i} - \mu_{f_y^k}} \right)^{\top} \in %\mathbb{R}^{d \times d}}.
%\end{equation}

\textbf{3.3.3 Feature Sampling and Label Assignment:} To enrich the diversity of new samples, Monte Carlo sampling is applied to generate new sample features from the normal distribution $N(\mu_{f_y^k},\Sigma_{f_y^k})$, as shown in Fig. \ref{fig: TNS}(c). After sampling, each new sample feature needs to be assigned a reasonable label. It is compared with each original sample feature in the slice, and assigned the label of the most similar original sample. It will be inserted before or after the timepoint of the original one to generate a new time-series feature.
\subsection{Convolution Weighted Loss}
Figure \ref{fig:method}(d) shows that the required sample quantities at some regression labels are less than the original ones. However, undersampling will break the contextual continuity of samples. Inspired by Dense Loss \cite{steininger2021density} and LDS \cite{yang2021delving}, we reweight these samples using the ratio of sample quantities before and after convolution at each label.

Let $n_{y}$ and $\Tilde{n}_{y}$ denote the sample quantities of label $y$ before and after convolution, respectively. For sample $x_i$ with label $y_i$, its normalized weight is
% \begin{equation}
%     w_{y_i} = \frac{{\overset{\sim}{n}}_{y_i}}{n_{y_i}}.
% \end{equation}
\begin{equation}
    w_{{y_i}} = \frac{m ({\overset{\sim}{n}}_{y_i} / n_{y_i})}{\sum_{j = 1}^{m}({\overset{\sim}{n}}_{y_j} / n_{y_j})}.
\end{equation}
Then, Convolution Weighted Loss (CWL) is
\begin{equation}
    Loss_{CW} = \frac{1}{m}{\sum_{i = 1}^{m}{{w_{y_i}}\left( {y_{i} - {\hat{y}}_{i}} \right)^{2},}}
\end{equation}
where $m$ is the number of frames in a time-series data, $y_i$ and $\hat{y}_i$ are the ground truth and the predicted label for $x_i$, respectively. CWL can maintain the contextual continuity of time-series data by lowering sample weight rather than eliminating samples. Unlike Inverse Frequency Loss \cite{huang2016learning} and Dense Loss \cite{steininger2021density} expecting models to be trained on a balanced label distribution, CWL is based on an approximated utopia label distribution.
\section{Experiments}
To evaluate the performance of our proposed method, we conducted experiments on three benchmark datasets for STR tasks, including movie affective analysis and video summarization.

\subsection{Datasets and Evaluation Metrics.}
\textbf{LIRIS-ACCEDE} \cite{baveye2015liris} is a widely used corpus of video content for movie affective analysis, comprising 160 movies with continuous valence and arousal scores ranging from -1 to 1 at every second along the movie. The dataset includes a variety of themes such as horror, comedy, and action, among others, and is available in multiple languages, including English, French, and Spanish.

\textbf{SumMe} \cite{gygli2014creating} consists of 25 user videos covering diverse events, such as cooking, sports, etc. The videos’ length ranges from 1.5 to 6.5 minutes. The annotation is the importance score for each frame, which was voted by 15 to 18 persons.

\textbf{TVSum} \cite{song2015tvsum} includes 50 videos collected from YouTube with 10 different categories, such as animal grooming, making sandwiches, changing vehicle tires, etc. The videos vary from 1 to 5 minutes in length and were annotated with a sequence of frame-level importance scores by 20 users.

For LIRIS-ACCEDE, the evaluation metrics are the Mean Square Error (MSE) and Pearson’s Correlation Coefficient (PCC). For SumMe and TVSum, we follow \cite{apostolidis2021combining} to employ F1-score as the evaluation metric.

\subsection{Baselines}
We evaluated our ULDA on three models for Subjective Time-series Regression tasks.

\textbf{PGL-SUM} \cite{apostolidis2021combining} \footnote{https://github.com/e-apostolidis/PGL-SUM.} was devised for video summarization tasks and has shown excellent performance on the SumMe and TVSum datasets. It segments each video into $M$ slices and models the dependencies between global and local frames using multi-head attention mechanisms. PGL-SUM combines the original deep representations with the representations encoding global and local dependencies, then passes them through dropout and normalization to a Regressor Network for score prediction.

\textbf{Encoder-only Transformer (ET)} \cite{park2022towards}\footnote{https://github.com/EIHW/MuSe2022} was the 3rd placed model in the MuSe-Stress 2022 challenge. The challenge aims at building sequence regression models for predicting valence and physiological arousal levels of people facing stressful conditions. ET consists of two linear layers, a position encoder, and a transformer encoder.

\textbf{RMN} \cite{zhang2022enlarging} was proposed for movie affective analysis and reportedly the best model on the LIRIS-ACCEDE dataset so far. It also showed a good generalization on other long sequence prediction tasks. RMN introduces a readable and writable memory bank to store useful historical features and employs a reinforcement learning scheme to update the content in the memory bank. We reproduced the RMN model based on the description in the paper.

\subsection{Implementation Details}
For LIRIS-ACCEDE, multi-modal features are extracted for each frame, including audio, background music, visual scene, human action, and facial expression. For audio features, a 128-dimensional vector is retrieved using VGGish \cite{hershey2017cnn} pre-trained on AudioSet \cite{gemmeke2017audio}. The background music (bgm) feature is a 128-dimensional vector extracted by a pre-trained VGGish. For scene features, a global max-pooling operation is performed on the last convolution layer of a VGG16 model that has been pre-trained on Places365 \cite{zhou2017places}, resulting in a 512-dimensional vector. For human action, two groups of convolution/max-pooling layers and a dense layer are appended to OpenPose's backbone \cite{cao2021openpose} and then fine-tuned on the LIRIS-ACCEDE dataset. A 128-dimensional action feature is extracted from the last dense layer. For human expression features, the largest face detected by MTCNN \cite{zhang2016joint} is utilized, and a 3072-dimensional vector is extracted using an Xception network \cite{chollet2017xception} that has been pre-trained on RAF \cite{li2017reliable}. To combine these frame-level features into a slice-level vector, a two-layer bidirectional LSTM with 128 units is employed. For the SumMe and TVSum datasets, the output of the penultimate layer (pool5) of GoogleNet \cite{szegedy2015going} pretrained on ImageNet \cite{russakovsky2015imagenet}, is used as the frame representation.

For the LIRIS-ACCEDE dataset, $b$ is set to 100, and for the SumMe and TVSum datasets, $b$ is set to 10. The kernel sizes $\delta$ and standard deviations $\sigma$ of Gaussian kernel functions $\mathbb{K(\cdot)}$ for convolution are 0.06, 0.3, 0.3 and 0.02, 0.1, 0.1 for LIRIS-ACCEDE, SumMe, and TVSum, respectively. When estimating the normal distribution of the feature set in a slice, the length threshold $T$ of the time slice is set to 10 for all datasets. As TNS generates new samples in feature space, we put it to the suitable place in each baseline according to their own architectures. For PGL-SUM, TNS is placed after the fusion of global and local features. For ET, TNS is placed after the Transformer encoder. TNS is not applied to RMN because it is progressively updated by samples along the time axis. Therefore, only CWL is applied to RMN. The hyperparameters of all three baselines are the same as the settings in their papers. More details can be found in the supplementary material.

\subsection{Main Results}
% \usepackage{multirow}

\begin{table}[]
\caption{Comparisons with SOTA methods on LIRIS-ACCEDE. The performances of competing methods are from \cite{zhang2022enlarging}. Three baselines are grouped into three subsections, whose results are from our implementations.  CWL stands for using only the Convolution Weighted Loss. TNS+CWL stands for using both the Time-slice Gaussian Sampling and the Convolution Weighted Loss. The overall best results are highlighted in red bold, while the best results for each subsection are in black bold.}
\label{tab:liris}
% \resizebox{\linewidth}{!}{
% \tiny
% \setlength{\tabcolsep}{1mm}{
\begin{tabular}{lllll}
\hline
                   & \multicolumn{2}{l}{\textbf{Valence}}                                          & \multicolumn{2}{l}{\textbf{Arosual}}                                          \\ \cline{2-5}
 & \textbf{MSE$\downarrow$}                          & \textbf{PCC$\uparrow$}                          & \textbf{MSE$\downarrow$}                          & \textbf{PCC$\uparrow$}                          \\ \hline
\multicolumn{5}{l}{\textbf{Method Comparison}}                                                                                                                                     \\
Yi et al. \cite{yi2018cnn}          & 0.090                                 & 0.301                                 & 0.136                                 & 0.175                                 \\
GLA  \cite{sun2019gla}              & 0.084                                 & 0.278                                 & 0.133                                 & 0.351                                 \\
Ko et al. \cite{ko2018towards}         & 0.102                                 & 0.114                                 & 0.149                                 & 0.083                                 \\
Zhao et al. \cite{zhao2019video}      & 0.071                                 & 0.444                                 & 0.137                                 & 0.419                                 \\
Affect2MM   \cite{mittal2021affect2mm}       & {\textcolor{red}{\textbf{0.068}}} & \quad-                                     & 0.128                                 & \quad-                                     \\
RMN \cite{zhang2022enlarging}               & {\textcolor{red}{\textbf{0.068}}} & 0.471                                 & 0.124                                 & 0.468                                 \\
RMN+ULDA(Ours)         & 0.069                                 & {\textcolor{red}{\textbf{0.531}}} & {\textcolor{red}{\textbf{0.114}}} & {\textcolor{red}{\textbf{0.492}}} \\ \hline
\multicolumn{5}{l}{\textbf{ET \cite{park2022towards}}}                                                                                                                                                    \\
Baseline           & 0.092                                 & 0.358                                 & 0.134                                 & 0.308                                 \\
CWL                & \textbf{0.079}                        & 0.429                                 & 0.131                                 & 0.334                                 \\
TNS+CWL            & 0.082                                 & \textbf{0.466}                        & \textbf{0.129}                        & \textbf{0.394}                        \\ \hline
\multicolumn{5}{l}{\textbf{PGL-SUM \cite{apostolidis2021combining}}}                                                                                                                                               \\
Baseline           & 0.090                                 & 0.429                                 & 0.136                                 & 0.441                                 \\
CWL                & 0.076                                 & 0.460                                 & 0.124                                 & 0.432                                 \\
TNS+CWL            & \textbf{0.069}                        & \textbf{0.475}                        & \textbf{0.120}                        & \textbf{0.445}                        \\ \hline
\multicolumn{5}{l}{\textbf{RMN \cite{zhang2022enlarging}}}                                                                                                                                                   \\
Baseline           & 0.073                                 & 0.515                                 & 0.124                                 & 0.441                                 \\
CWL                & \textbf{0.069}                        & {\textcolor{red}{\textbf{0.531}}} & {\textcolor{red}{\textbf{0.114}}} & {\textcolor{red}{\textbf{0.492}}} \\ \hline
\end{tabular}
% }
\end{table}
% \vspace{-0.5 cm}

\begin{table}[t]
\caption{Comparisons with SOTA methods on SumMe and TVSum. The performances of competing methods are from \cite{zhang2022enlarging}. }% Three baselines are grouped into three subsections, whose results are from our implementations.  CWL stands for using only the Convolution Weighted Loss. TNS+CWL stands for using both the Time-slice Gaussian Sampling and the Convolution Weighted Loss. The overall best results are highlighted in red bold. The best results for each subsection are in black bold.}
\label{tab:sum}
\setlength{\tabcolsep}{5.5mm}{
% \resizebox{\linewidth}{!}{
% \small
% \fontsize{8.3pt}{\baselineskip}\selectfont
\begin{tabular}{lcc}
\bottomrule
              & \multicolumn{2}{c}{\textbf{F1-score$\uparrow$}}                                                \\ \cline{2-3}
              & \textbf{SumMe}                                & \textbf{TVSum}                                \\ \bottomrule
\multicolumn{3}{l}{\textbf{Method Comparison}}                                              \\
A-AVS \cite{ji2019video}         & 43.9                                 & 59.4                                 \\
M-AVS \cite{ji2019video}         & 44.4                                 & 61.0                                 \\
DASP \cite{ji2020deep}        & 45.5                                 & 63.6                                 \\
CRSum \cite{yuan2019spatiotemporal}        & 47.3                                 & 58.0                                 \\
MAVS  \cite{feng2018extractive}        & 40.3                                 & 66.8                                 \\
SUM-GDA \cite{li2021exploring}      & 52.8                                 & 58.9                                 \\
SMLD \cite{chu2019spatiotemporal}         & 47.6                                 & 61.0                                 \\
H-MAN \cite{liu2019learning}        & 51.8                                 & 60.4                                 \\
SMN  \cite{wang2019stacked}         & 58.3                                 & 64.5                                 \\
PGL-SUM \cite{apostolidis2021combining}      & 55.6                                 & 61.0                                 \\
MSVA  \cite{ghauri2021supervised}        & 54.5                                 & 62.8                                 \\
RMN \cite{zhang2022enlarging}          & 63.0                                 & {\textcolor{red}{\textbf{67.6}}} \\
RMN+ULDA   (Ours) & {\textcolor{red}{\textbf{65.8}}} & 67.3                                 \\ \bottomrule
\multicolumn{3}{l}{\textbf{ET \cite{park2022towards}}}                                                             \\
Baseline      & 52.7                                 & 61.6                                 \\
CWL           & 54.8                                 & 62.9                           \\
TNS + CWL     & \textbf{56.7}                        & \textbf{66.0}                  \\ \bottomrule
\multicolumn{3}{l}{\textbf{PGL-SUM \cite{apostolidis2021combining}}}                                                        \\
Baseline      & 54.6                                 & 60.8                                 \\
CWL           & 57.9                                 & 62.1                          \\
TNS + CWL     & \textbf{58.3}                  & \textbf{63.3}                 \\ \bottomrule
\multicolumn{3}{l}{\textbf{RMN \cite{zhang2022enlarging}}}                                                            \\
Baseline      & 64.0                                 & 66.2                                 \\
CWL           & {\textcolor{red}{\textbf{65.8}}}                  & \textbf{67.3}                  \\ \bottomrule
\end{tabular}
}
\end{table}

We conducted the evaluation of our ULDA on three baselines on three subjective time-series regression datasets. For each dataset, we first compare RMN+ULDA with other competing methods, then group three baselines into three subsections to demonstrate the performance gains contributed by ULDA for each baseline. Each test was performed three times, and the average result is reported in Tables \ref{tab:liris} and \ref{tab:sum}. Baseline denotes the result of the model we reproduced, CWL stands for the results of using only the CWL after convolving the original label distribution, and TNS + CWL stands for the result of using TNS for labels that need to be oversampled and CWL for labels that need to be undersampled after the label convolution.

The results of video emotion prediction on the LIRIS-ACCEDE dataset are shown in Table \ref{tab:liris}. Our best model, RMN+ULDA, outperforms almost all competing methods. We can see that the latest SOTA method, RMN, has been considerably improved on PCC of Valence and MSE and PCC of Arousal after being trained on the approximated utopia label distribution. Only the MSE of valence remained nearly unchanged. We also evaluate the effectiveness of CWL and TNS on each baseline. Please note that we implemented three baselines and tested them in our computing environment, and there existed a performance gap between our implementation and the one reported in \cite{zhang2022enlarging}. One can see CWL alone can improve their performance. For MSE, CWL relatively improves baselines by about 11.7\% and 6.4\% on average for valence and arousal, respectively. For PCC, CWL relatively improves baselines by about 10.0\% and 6.0\% on average, respectively. After combining CWL and TNS, the performance of baselines was further boosted. For MSE, TNS + CWL relatively improves baselines on valence and arousal by about 17.1\% and 7.8\% on average, respectively. For PCC, it relatively improves baselines by about 20.5\% and 14.4\% on average, respectively. We believe the reason behind this additional performance gain is as follows. The data space cannot be well represented by insufficient samples at rare labels. In contrast, TNS increases data diversity by generating new samples via Monte Carlo sampling, allowing models to learn a better data representation.

The results of video summarization on the SumMe and TVSum datasets are shown in Table 2. We can observe similar results. Our best model, RMN+ULDA, significantly outperforms all competing methods on the SumMe dataset and achieves a similar F1-score compared with RMN on the TVSum dataset. For baselines, TNS + CWL makes absolute improvements of about 3.9\% and 3.5\% on average on SumMe and TVSum, respectively. All of the above results demonstrate that the approximated utopia label distribution is more effective than the original label distribution on STR tasks.
\begin{figure*}
    \centering
    \includegraphics[width= 0.9\linewidth]{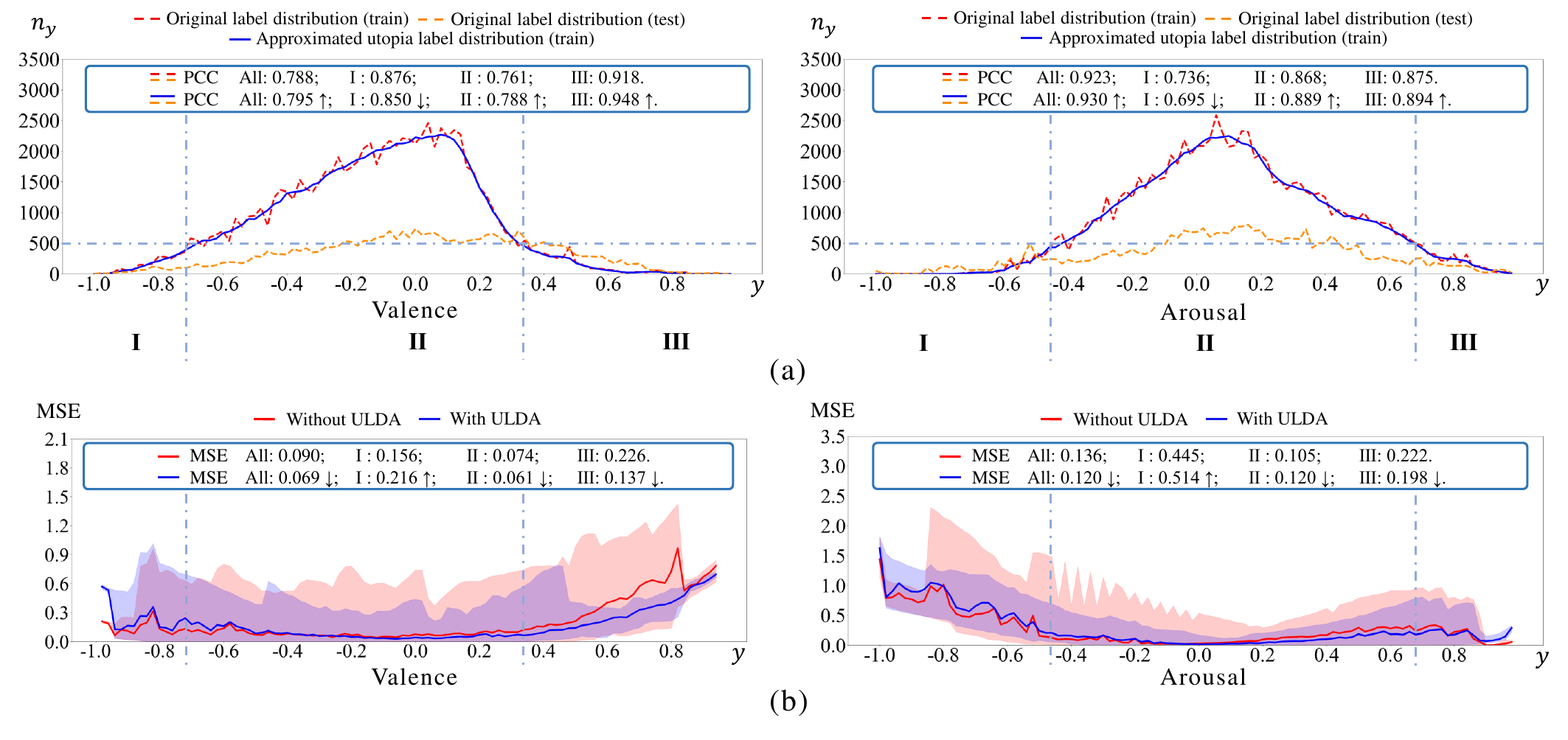}
    % \vspace{-0.2 cm}
    \caption{Correlation of performance and label distribution. (a) The label distributions of the training set with and without Gaussian convolution and the label distribution of the test set on the LIRIS-ACCEDE dataset. They are in blue, red, and yellow, respectively. And their PCC values. (b) MSE results of the PGL-SUM model trained with and without ULDA. The solid lines and the shadow ribbons represent the mean and range of MSEs for all samples at each label, respectively. And the mean MSEs in three regions.}
    \label{fig: corr}
    % \vspace{-0.2 cm}
\end{figure*}

\begin{table}[t]
\caption{Evaluation of varied losses on the LIRIS-ACCEDE dataset. The best results for each baseline are highlighted in bold, while the second best are italicized.}
\label{tab:loss}
% \resizebox{\linewidth}{!}{
% \tiny
% \fontsize{4pt}{5pt}\selectfont
\setlength{\tabcolsep}{3.5mm}{
\begin{tabular}{lllll}
\hline
           & \multicolumn{2}{c}{\textbf{Valence}} & \multicolumn{2}{c}{\textbf{Arousal}} \\ \cline{2-5}
           & \textbf{MSE$\downarrow$}      & \textbf{PCC$\uparrow$}     & \textbf{MSE$\downarrow$}      & \textbf{PCC$\uparrow$}     \\ \hline
\multicolumn{5}{l}{\textbf{ET \cite{park2022towards}}}                                                          \\
Baseline   & 0.092             & 0.358            & \textit{0.134}             & \textit{0.308}            \\
INV Loss   & 0.085             & 0.329            & 0.141             & 0.282            \\
LDS        & \textit{0.083}             & 0.348            & 0.138             & 0.298            \\
Dense Loss & 0.093             & \textit{0.387}            & 0.136             & \textbf{0.334}   \\
CWL        & \textbf{0.079}    & \textbf{0.429}   & \textbf{0.131}    & \textbf{0.334}   \\ \hline
\multicolumn{5}{l}{\textbf{PGL-SUM \cite{apostolidis2021combining}}}                                                     \\
Baseline   & 0.090             & 0.429            & 0.136             & \textit{0.441}            \\
INV Loss   & 0.082             & \textit{0.450}            & 0.135             & \textbf{0.452}   \\
LDS        & \textit{0.078}             & 0.437            & \textit{0.129}             & 0.415            \\
Dense Loss & 0.079             & 0.421            & 0.132             & 0.423            \\
CWL        & \textbf{0.076}    & \textbf{0.460}   & \textbf{0.124}    & 0.432            \\ \hline
\multicolumn{5}{l}{\textbf{RMN \cite{zhang2022enlarging}}}                                                         \\
Baseline   & 0.073             & 0.515            & \textit{0.124}             & 0.441            \\
INV Loss   & 0.073             & 0.502            & \textit{0.124}             & 0.434            \\
LDS        & \textit{0.072}             & \textit{0.521}            & 0.126             & \textit{0.446}            \\
Dense Loss & \textbf{0.069}    & 0.438            & 0.128             & 0.411            \\
CWL        & \textbf{0.069}    & \textbf{0.531}   & \textbf{0.114}    & \textbf{0.492}   \\ \hline
\end{tabular}
}
\end{table}

\subsection{Correlation Analysis of Performance and Label Distribution}
To have an insight into performance improvement, we performed a correlation analysis between the label distributions of the training set with and without Gaussian convolution and the label distribution of the test set on the LIRIS-ACCEDE dataset, as shown in Fig. \ref{fig: corr}(a). They are in blue, red, and yellow, respectively. In this analysis, we use the PGL-SUM model as an example. The MSE results of PGL-SUM trained with and without ULDA are plotted in Fig. \ref{fig: corr}(b). The solid lines and the shadow ribbons denote the mean MSE and the range of MSEs of all samples at each label, respectively. Overall, we can see that the PCC of the approximated utopia label distribution of the training set and the label distribution of the test set has been increased. In other words, these two label distributions become more similar to each other after convolution. At the same time, the mean MSE is decreased (the blue lines in Fig. \ref{fig: corr}(b)). More importantly, the range of MSEs becomes much narrower (the blue ribbons).

We observe that the mean MSE varies with the sample quantity. Therefore, the label distribution of the training set is further divided into one region, II, with high sample quantities, and two regions, I and III, with low sample quantities using 500 as the threshold. In region II, the mean MSE is small, whether using ULDA or not. In region III, the mean MSE decreases, while in region I, the mean MSE increases. One can see that the change of the mean MSE and the change of PCC are negatively correlated. However, the range of MSEs narrows across almost three regions, indicating improved prediction stability (model fairness).

Based on the observations above, we can draw two conclusions: 1) The performance gain is primarily due to the approximated utopia label distribution being closer to the label distribution of the test set; 2) Models can produce more stable predictions after being trained on the approximated utopia label distribution.

\subsection{Ablation Studies}
We conducted ablation studies on the LIRIS-ACCEDE dataset to analyze the effectiveness of various weighted losses and oversampling methods, as well as the key parameters in ULDA.

\textbf{4.6.1 Weighted Losses: }
To test the effectiveness of Convolution Weighted Loss (CWL), we compared it with three other losses for imbalance regression. They are: 1) Inverse-frequency weighting (INV) \cite{huang2016learning}, which reweights samples at each label based on the multiplicative inverse of the observed probability density of the label; 2) Label distribution smoothing (LDS) \cite{yang2021delving}, which first smooths the label distribution by a Gaussian kernel to estimate the label density distribution including label correlation, and then apply the INV loss; 3) Dense Loss \cite{steininger2021density}, which estimates the label density distribution similarly as LDS does, and then reweights samples using the additive inverse of the smoothed label probability density distribution. The assumption of these losses makes them different from CWL. They assume the test set is uniformly distributed, and expect models to be trained on a balanced training set. Nevertheless, CWL expects models to learn on the utopia label distribution, not necessarily uniform but closer to the real world.

The results in Table \ref{tab:loss} show that CWL performs the best in almost all tests. This suggests that the assumption of a uniformly distributed dataset may not hold true for STR tasks, and confirms the validity of our proposed approximated utopia Label distribution.

\begin{table}[t]
\caption{Evaluation of varied oversampling methods on the LIRIS-ACCEDE dataset.}
\label{tab:augment}
% \resizebox{\linewidth}{!}{
% \tiny
% \fontsize{4pt}{5pt}\selectfont
\setlength{\tabcolsep}{3.5mm}{
\begin{tabular}{lllll}
\hline
         & \multicolumn{2}{c}{\textbf{Valence}} & \multicolumn{2}{c}{\textbf{Arousal}} \\ \cline{2-5}
         & \textbf{MSE$\downarrow$}      & \textbf{PCC$\uparrow$}     & \textbf{MSE$\downarrow$}      & \textbf{PCC$\uparrow$}     \\ \hline
\multicolumn{5}{l}{\textbf{ET \cite{park2022towards}}}                                                        \\
Baseline & 0.092             & 0.358            & 0.134             & 0.308            \\
SMOGN    & \textit{0.086}             & \textit{0.424}            & \textbf{0.129}    & 0.296            \\
C-Mixup  & 0.087             & 0.371            & \textit{0.131}             & \textit{0.353}            \\
TNS      & \textbf{0.082}    & \textbf{0.466}   & \textbf{0.129}    & \textbf{0.394}   \\ \hline
\multicolumn{5}{l}{\textbf{PGL-SUM \cite{apostolidis2021combining}}}                                                   \\
Baseline & 0.090             & 0.429            & 0.136             & \textit{0.441}            \\
SMOGN    & \textbf{0.066}    & \textbf{0.508}   & \textit{0.129}             & 0.394            \\
C-Mixup  & 0.078             & 0.445            & 0.134             & 0.410            \\
TNS      & \textit{0.069}             & \textit{0.475}            & \textbf{0.120}    & \textbf{0.445}   \\ \hline
\end{tabular}
}
\end{table}
\textbf{4.6.2 Oversampling Methods:}
To evaluate TNS, we compared it with two well-accepted oversampling methods for regression tasks: 1) SMOGN \cite{branco2017smogn}, which combines SmoteR \cite{torgo2013smote} with the Gaussian noise to generate new samples; 2) C-Mixup \cite{zhang2017mixup}, which selects more reliable neighboring samples based on label distances for new sample synthesis. In this test, SMOGN and C-Mixup only select samples in the slice for oversampling. This can maintain the continuity of new samples in temporal context space.

To fit the approximated utopia label distribution, we used CWL to decrease the loss of samples in the labels that need to be undersampled.

The results in Table \ref{tab:augment} show that TNS achieved the best performance on the ET model. For the PGL-SUM model, TNS is the best for arousal prediction and the second-best for valence prediction. This demonstrates that TNS is more effective than other the two methods for STR tasks because it can maintain the contextual continuity of new samples.

\textbf{4.6.3 Key Parameters:}
ULDA has two key parameters. The kernel size $\delta$ and standard deviation $\sigma$ of the Gaussian kernel $\mathbb{k(\cdot)}$ for convoluting the label distribution. Here we use the PGL-SUM model as an example. Please refer to the supplementary material for the studies on the other two baselines. Table 5 shows the results of a controlled experiment by varying $\delta \in \{0.06,0.10,0.14\}$ and $\sigma \in \{0.02,0.04,0.06\}$ alternatively. We can see the best setting is $\delta=0.06$ and $\sigma=0.02$.

ULDA has another key parameter. The threshold of time slice length $T$ in TNS to generate new samples. The results in Table 6 show that all settings can improve the performance of the model. Considering the overall performance and the computational cost, $T$ is set as 10.

% \usepackage{multirow}
\begin{table}[t]
\caption{Evaluation of varied kernel size $\delta$ and standard deviation $\sigma$.}
\label{tab:kernel}
% \resizebox{\linewidth}{!}{
% \tiny
% \fontsize{4pt}{5pt}\selectfont
\setlength{\tabcolsep}{3.0mm}{
\begin{tabular}{llllll}
\hline
\multicolumn{2}{l}{} &
\multicolumn{2}{c}{\textbf{Valence}} &
\multicolumn{2}{c}{\textbf{Arousal}} \\ \cline{3-6}
\multicolumn{2}{l}{}                  & \textbf{MSE$\downarrow$}      & \textbf{PCC$\uparrow$}     & \textbf{MSE$\downarrow$}      & \textbf{PCC$\uparrow$}     \\ \hline
\multicolumn{2}{l}{\textbf{w/o ULDA}}  & 0.090             & 0.429            & 0.136             & 0.441            \\
$\delta$                & $\sigma$                &                   &                  &                   &                  \\ \hline
0.06                 & 0.02                 & \textbf{0.069}    & \textbf{0.475}   & \textbf{0.120}    & \textbf{0.445}   \\
0.06                 & 0.04                 & 0.074             & \textit{0.473}            & 0.133             & 0.413            \\
0.06                 & 0.06                 & \textbf{0.069}    & 0.454            & 0.133             & 0.413            \\
0.10                 & 0.02                 & \textit{0.071}             & 0.445            & 0.128             & 0.405            \\
0.10                 & 0.04                 & 0.073             & 0.460            & 0.130             & 0.416            \\
0.10                 & 0.06                 & 0.074             & 0.462            & 0.130             & 0.420            \\
0.14                 & 0.02                 & \textbf{0.069}    & 0.471            & 0.132             & 0.420            \\
0.14                 & 0.04                 & 0.072             & 0.437            & 0.128             & 0.403            \\
0.14                 & 0.06                 & \textbf{0.069}    & 0.462            & \textit{0.126}             & \textit{0.436}            \\ \hline
\end{tabular}
}
\end{table}

% \usepackage{multirow}
\begin{table}[t]
\caption{Evaluation of varied time slice length threshold $T$.}
\label{tab:T}
% \resizebox{\linewidth}{!}{
% \tiny
% \fontsize{4pt}{5pt}\selectfont
\setlength{\tabcolsep}{3.0mm}{
\begin{tabular}{lllll}
\hline
         & \multicolumn{2}{c}{\textbf{Valence}} & \multicolumn{2}{c}{\textbf{Arousal}} \\ \cline{2-5}
                  & \textbf{MSE$\downarrow$}      & \textbf{PCC$\uparrow$}     & \textbf{MSE$\downarrow$}      & \textbf{PCC$\uparrow$}     \\ \hline
\textbf{w/o ULDA}  & 0.090             & 0.429            & 0.136             & 0.441            \\
\multicolumn{5}{l}{$T$}                                                                           \\ \hline
5                 & \textbf{0.065}    & 0.453            & \textit{0.130}             & 0.405            \\
10                & 0.069             & \textit{0.475}            & \textbf{0.120}    & \textbf{0.445}   \\
15                & 0.072             & \textbf{0.524}   & \textit{0.130}             & \textit{0.443}            \\
20                & \textit{0.066}             & 0.460            & \textit{0.130}             & 0.417            \\ \hline
\end{tabular}
}
\end{table}

\section{conclusion}
We introduce a novel Utopia Label Distribution Approximation for the label distribution bias problem of time-series data in STR tasks. ULDA improves the model fairness by making label distributions of the training and test sets more similar to each other. As a result, ULDA boosts conventional methods for STR tasks to achieve new state-of-the-art performances on three benchmarks. We hope that our proposal will spark further research into the training of models with biased label distribution.

\textbf{Limitations and future work.} The analysis in section 4.5 shows ULDA cannot always increase the PCC between the approximated utopia label distribution and the label distribution of the test set, especially for some rare labels. A possible reason is that those sampled test data differ greatly from the real world. We will investigate more advanced solutions in future work. Furthermore, the current TNS is inapplicable to the model that is progressively updated by samples along the time axis. This is left for a further extension to allow TNS to be model agnostic.

\vfill\pagebreak

% \section{REFERENCES}
% \label{sec:refs}

% List and number all bibliographical references at the end of the
% paper. The references can be numbered in alphabetic order or in
% order of appearance in the document. When referring to them in
% the text, type the corresponding reference number in square
% brackets as shown at the end of this sentence \cite{C2}. An
% additional final page (the fifth page, in most cases) is
% allowed, but must contain only references to the prior
% literature.

\bibliographystyle{IEEEbib}
\bibliography{strings,refs}

% \vfill\pagebreak
\section{appendix}
\subsection{Theoretical Analysis}
The theory behind approximating the utopia label distribution by a Gaussian convolution is the central limit theorem (CLT). In probability theory, CLT establishes that, in many situations, for identically distributed independent samples, the distribution of a sample variable approximates a normal distribution as long as the sample size is large enough, even if the original variables themselves are not normally distributed. Let’s consider the label as a variable of a sample. In probability theory, the central limit theorem tells us that the label mean tends towards a Gaussian distribution in the real world. Specifically, assume the label space of data has $b$ regression labels. Given a frame sample $x_i,(i=1,…,m)$ with true label $y$ in a time-series data, people label the sample according to their own cognition. Their annotations may not be consistent due to the independence and subjectivity of human cognition. The majority of people will label $x_i$ as $y$, while the minority will label it as neighboring labels $y \pm \sigma$. In existing datasets, $y$ is usually assigned by the average or majority voting of $n$ annotators because $n \ll \infty$. However, in the real world, when $n \approx \infty$, the label mean of the sample tends towards a Gaussian distribution $N(y,\sigma _i^2)$ with mean $y$ and variance $\sigma_i^2$, where $\sigma_i^2$ denotes the variance of people’s inconsistent votes. This means that each sample can be labeled as $y$ with a possibility following $N(y,\sigma _i^2)$. When the population of samples with the true label $y$ is increased to $n_{y}$, they would be labeled as $y$ with a possibility of following a Gaussian distribution $N(y,\sigma _{y}^2)$, where
\begin{equation}
\mu_{y}=\frac{\sum_{i=1}^{n_{y}} y}{{n_{y}}}=y,\qquad  \sigma_{y}=\frac{\sum_{i=1}^{n_{y}}\sigma_i^2}{n^2_{y}}.
\end{equation}
From another perspective, the label distribution of this population follows $N(y,\sigma^2_{y})$ as well. We can see the utopia sample quantity at each regression label can be estimated by the Gaussian weighted sum of the sample quantity at the label and sample quantities at neighboring labels in the database.

\subsection{The experiment on music data}

\begin{table}[t]
\caption{Comparisons with SOTA methods on PMEmo.}
\label{tab:PMEmo}
% \resizebox{\linewidth}{!}{
% \tiny
\begin{tabular}{lllll}
\hline
                   & \multicolumn{2}{c}{\textbf{Valence}}                                          & \multicolumn{2}{c}{\textbf{Arosual}}                                              \\ \cline{2-5}
& \textbf{MSE$\downarrow$}                          & \textbf{PCC$\uparrow$}                          & \textbf{MSE$\downarrow$}                          & \textbf{PCC$\uparrow$}                          \\ \hline
\multicolumn{5}{l}{\textbf{Method Comparison}}                                                                                                                                     \\
Zhang et al. [49]       & 0.088                                 & 0.095                                 & {\textcolor{red}{\textbf{0.085}}} & 0.115                                 \\
RMN [48]                & {\textcolor{red}{\textbf{0.087}}} & 0.206                                 & 0.089                                 & 0.236                                 \\
RMN+ULDA(Ours)     & {\textcolor{red}{\textbf{0.087}}} & 0.251                                 & 0.088                                 & 0.261                                 \\ \hline
\multicolumn{5}{l}{\textbf{ET [28]}}                                                                                                                                                    \\
Baseline           & 0.106                                 & 0.117                                 & 0.096                                 & 0.121                                 \\
CWL                & 0.104                                 & 0.117                                 & \textbf{0.092}                        & 0.145                                 \\
TNS+CWL            & \textbf{0.104}                        & \textbf{0.130}                        & \textbf{0.092}                        & \textbf{0.164}                        \\ \hline
\multicolumn{5}{l}{\textbf{PGL-SUM [1]}}                                                                                                                                               \\
Baseline           & 0.103                                 & 0.251                                 & 0.104                                 & 0.268                                 \\
CWL                & 0.102                                 & {\textcolor{red}{\textbf{0.304}}} & 0.100                                 & 0.276                                 \\
TNS+CWL            & \textbf{0.100}                        & 0.280                                 & \textbf{0.098}                        & {\textcolor{red}{\textbf{0.293}}} \\ \hline
\multicolumn{5}{l}{\textbf{RMN [48]}}                                                                                                                                                   \\
Baseline           & 0.089                                 & 0.195                                 & 0.089                                 & 0.236                                 \\
CWL                & {\textcolor{red}{\textbf{0.087}}} & \textbf{0.251}                        & \textbf{0.088}                        & \textbf{0.261}                        \\ \hline
\end{tabular}
% }
\end{table}

% \usepackage{multirow}
\begin{table}[h]
\caption{Evaluation of varied kernel size $\delta$ and standard deviation $\sigma$.}
\label{tab:supp_kernel}
% \resizebox{\linewidth}{!}{
% \tiny
% \fontsize{4pt}{5pt}\selectfont
\begin{tabular}{llllll}
\hline
\multicolumn{2}{l}{} &
\multicolumn{2}{c}{\textbf{Valence}} &
\multicolumn{2}{c}{\textbf{Arousal}} \\ \cline{3-6}
\multicolumn{2}{l}{}                        & \textbf{MSE$\downarrow$}      & \textbf{PCC$\uparrow$}     & \textbf{MSE$\downarrow$}      & \textbf{PCC$\uparrow$}     \\ \hline
\multicolumn{2}{l}{\textbf{w/o ULDA (ET)}}  & 0.092             & 0.358            & 0.134             & 0.308            \\
$\delta$           & $\sigma$         &                   &                  &                   &                  \\ \hline
0.06                     & 0.02                   & \textbf{0.082}    & \textbf{0.466}   & 0.129             & \textbf{0.394}   \\
0.06                     & 0.04                   & 0.088             & \textit{0.388}   & 0.137             & 0.316            \\
0.06                     & 0.06                   & 0.088             & \textit{0.388}   & 0.132             & 0.338            \\
0.10                     & 0.02                   & 0.095             & 0.344            & 0.138             & 0.309            \\
0.10                     & 0.04                   & 0.087             & 0.353            & 0.128             & \textit{0.353}   \\
0.10                     & 0.06                   & 0.095             & 0.355            & \textit{0.126}    & 0.324            \\
0.14                     & 0.02                   & 0.088             & 0.342            & 0.139             & 0.300            \\
0.14                     & 0.04                   & \textit{0.086}    & 0.380            & \textbf{0.123}    & 0.341            \\
0.14                     & 0.06                   & 0.090             & 0.369            & 0.133             & 0.334            \\ \hline
\multicolumn{2}{l}{\textbf{w/o ULDA (RMN)}} & 0.073             & 0.515            & 0.124             & 0.441            \\
$\delta$           & $\sigma$         &                   &                  &                   &                  \\ \hline
0.06                     & 0.02                   & \textbf{0.069}    & \textbf{0.531}   & \textbf{0.114}    & \textbf{0.492}   \\
0.06                     & 0.04                   & 0.072             & 0.522            & 0.122             & \textit{0.491}   \\
0.06                     & 0.06                   & 0.072             & 0.522            & 0.122             & \textit{0.491}   \\
0.10                     & 0.02                   & \textit{0.070}    & 0.520            & 0.124             & 0.470            \\
0.10                     & 0.04                   & 0.075             & 0.504            & 0.129             & 0.443            \\
0.10                     & 0.06                   & \textbf{0.069}    & 0.521            & 0.120             & 0.468            \\
0.14                     & 0.02                   & 0.074             & 0.473            & 0.118             & 0.467            \\
0.14                     & 0.04                   & 0.081             & 0.501            & \textit{0.116}    & 0.477            \\
0.14                     & 0.06                   & 0.073             & \textit{0.524}   & 0.125             & 0.463            \\ \hline
\end{tabular}
% }
\end{table}

\begin{table}[t]
\caption{Evaluation of varied time slice length threshold $T$ on ET.}
\label{tab:supp_T}
% \resizebox{\linewidth}{!}{
% \tiny
% \fontsize{4pt}{5pt}\selectfont
\setlength{\tabcolsep}{3.0mm}{
\begin{tabular}{lllll}
\hline
         & \multicolumn{2}{c}{\textbf{Valence}} & \multicolumn{2}{c}{\textbf{Arousal}} \\ \cline{2-5}
                  & \textbf{MSE$\downarrow$}      & \textbf{PCC$\uparrow$}     & \textbf{MSE$\downarrow$}      & \textbf{PCC$\uparrow$}     \\ \hline
\textbf{w/o ULDA}       & 0.092             & 0.358            & 0.134             & 0.308   \\
\multicolumn{5}{l}{$T$}                                                                           \\ \hline
5                 & 0.089             & 0.389            & 0.134             & 0.332            \\
10                & \textbf{0.082}    & \textbf{0.466}   & \textbf{0.129}    & \textbf{0.394}   \\
15                & 0.086             & \textit{0.439}   & \textit{0.131}    & \textit{0.349}   \\
20                & \textit{0.085}    & 0.379            & 0.138             & 0.280            \\ \hline
\end{tabular}
}
\end{table}

To further demonstrate the generalization of ULDA, we conducted experiments on the music emotion recognition dataset, PMEmo. The PMEmo dataset contains 794 music choruses with dynamic labels for each 0.5-second segment, where each label was annotated by at least 10 persons. The evaluation metrics are the Root Mean Square Error (RMSE) and Pearson’s Correlation Coefficient (PCC).
Table \ref{tab:PMEmo} shows the results on PMEmo, where we can observe similar results. RMN+ULDA significantly outperforms competing methods on PCC of both valence and arousal. For MSE, it achieves a similar result to those of the competing methods. For baselines, CWL alone can substantially improve the results of both MSE and PCC on valence and arousal. After combining CWL and TNS, the performance is further boosted in almost all indicators.

\subsection{Experimental Settings}
The detailed settings of the three baselines and how TNS is applied to PGL-SUM and ET are given in below.

\textbf{PGL-SUM}: The model consists of three main components: a video encoder, a self-attention module, and a regressor network. The original video is segmented into 4 slices and the self-attention module consists of 8 heads. TNS is applied on the feature set $f \in \mathbb{R}^{m \times d}$ that fuses local and global information, where $d$ is the dimension size set to 128 and $m$ is the length of the time-series data. The augmented time-series features are then fed into the regressor network for prediction.

The learning rate, dropout rate and L2 regularization factor are $5 \cdot 10^{-5}$, $0.5$, and $10^{-5}$, respectively. For the network's weights initialization, the Xavier uniform initialization approach with gain = $\sqrt{2}$ and biases = 0.1 is used. Training is performed in a full-batch mode (i.e., the batch size is equal to the number of samples) using Adam optimizer, and stops after 200 epochs.

\textbf{ET}: The model is an encoder-only transformer structure with a regressor network. The number of heads is set to 8. TNS is applied on the output feature set $f \in \mathbb{R}^{m \times d}$ of the transformer, where $d$ is the dimension size set to 128 and $m$ is the length of the time-series data. The augmented time-series features are then fed into the regressor network.

During training, the early stopping strategy is adopted and the number of maximum epochs is set to 200. Adam optimizer is used to optimize the model parameters and the learning rate is set to $10^{-4}$.

\textbf{RMN}: RMN consists of three main components: a feature extraction module, a memory bank, and a value network. The feature extraction module extracts features from every 10-second clip of the movie. The memory bank is a matrix with shape $N \times d$, where the dimension $d$ is 128 and the size N is set to 10. It stores historical information and selectively updates the contents every 32 steps via an RL-based update scheme. All parameters are optimized using Adam with a learning rate of $10^{-4}$.

\subsection{Ablation Studies}

We also conducted ablation studies on key parameters of the other two baselines. Table \ref{tab:supp_kernel} shows the results of $\delta$ and $\sigma$ on ET and RMN. Table \ref{tab:supp_T} shows the results of $T$ on ET since TNS is not applied to RMN. We can see the best setting is $\delta = 0.06$, $\sigma = 0.02$ and $T = 10$, which is consistent with the setting of PGL-SUM.

\end{document}